\documentclass{article}

\usepackage[preprint,nonatbib]{neurips_2024}

\usepackage{multirow}
\usepackage{arydshln}
\usepackage{wrapfig}
\usepackage[utf8]{inputenc} 
\usepackage{needspace}
\usepackage{float}
\usepackage{lipsum}
\usepackage[caption = true]{subfig}
\usepackage[T1]{fontenc}

\usepackage{hyperref}       
\usepackage{booktabs}       

\usepackage{nicefrac}       
\usepackage{microtype}      
\usepackage[table, svgnames, dvipsnames]{xcolor}       
\usepackage{pifont}
\usepackage{graphicx}
\usepackage{amsmath}
\usepackage{cleveref}
\usepackage{amsthm}
\usepackage{verbatim}
\usepackage{listings}
\usepackage{todonotes}
\usepackage{xcolor}

\usepackage{amssymb}
\usepackage{hyperref}
\hypersetup{colorlinks,linkcolor={red},citecolor={blue},urlcolor={red}} 
\usepackage{amsthm}

\usepackage{xcolor}

\definecolor{codegreen}{rgb}{0,0.6,0}
\definecolor{codegray}{rgb}{0.5,0.5,0.5}
\definecolor{codepurple}{rgb}{0.58,0,0.82}
\definecolor{backcolour}{rgb}{0.95,0.95,0.92}

\usepackage{listings}
\usepackage{color}

\definecolor{dkgreen}{rgb}{0,0.6,0}
\definecolor{gray}{rgb}{0.5,0.5,0.5}
\definecolor{mauve}{rgb}{0.58,0,0.82}

\title{\textbf{GPT-4o: Visual perception performance of }\textbf{multimodal large language models in piglet activity understanding}
}

\author{%
Yiqi Wu$^{1}$ \quad Xiaodan Hu $^{2}$ \quad Ziming Fu $^{3}$ \\\quad\textbf{Siling Zhou} $^1$ \quad\textbf{Jiangong Li }$^1$\thanks{Corresponding Author\\ jli153@cau.edu.cn (Jiangong Li)}\quad \\
$^1$State Key Laboratory of Animal Nutrition and Feeding, \\College of Animal Science and Technology, China Agricultural University, Beijing, China\\   \quad $^2$Department of Electrical and Computer Engineering, \\University of Illinois at Urbana-Champaign, Urbana, Illinois  \\\quad $^3$AniEye Technology, Hangzhou, Zhejiang, China \\
}
\begin{document}

\maketitle

\begin{abstract}

Animal ethology is an crucial aspect of animal research, and animal behavior labeling is the foundation for studying animal behavior. This process typically involves labeling video clips with behavioral semantic tags, a task that is complex, subjective, and multimodal. With the rapid development of multimodal large language models (LLMs), new application have emerged for animal behavior understanding tasks in livestock scenarios. This study evaluates the visual perception capabilities of multimodal LLMs in animal activity recognition. To achieve this, we created piglet test data comprising close-up video clips of individual piglets and annotated full-shot video clips. These data were used to assess the performance of four multimodal LLMs—Video-LLaMA, MiniGPT4-Video, Video-Chat2, and GPT-4 omni (GPT-4o)—in piglet activity understanding. Through comprehensive evaluation across five dimensions, including counting, actor referring, semantic correspondence, time perception, and robustness, we found that while current multimodal LLMs require improvement in semantic correspondence and time perception, they have initially demonstrated visual perception capabilities for animal activity recognition. Notably, GPT-4o showed outstanding performance, with Video-Chat2 and GPT-4o exhibiting significantly better semantic correspondence and time perception in close-up video clips compared to full-shot clips. The initial evaluation experiments in this study validate the potential of multimodal large language models in livestock scene video understanding and provide new directions and references for future research on animal behavior video understanding. Furthermore, by deeply exploring the influence of visual prompts on multimodal large language models, we expect to enhance the accuracy and efficiency of animal behavior recognition in livestock scenarios through human visual processing methods. 

\end{abstract}


\section{Introduction}

Understanding animal behavior is crucial for promoting animal welfare. Central to this understanding is the manual labeling of animal behavior, which is fundamental for analyzing these behaviors. A standard method of labeling animal postures and behaviors involves accurately describing animal behavior using semantic sentences and assigning behavioral labels to short video clips of animals, which is inherently complex, subjective, and multimodal.

Recent advancements in large language models (LLMs) have opened new opportunities for understanding and labeling animal behavior. Models like LLaMa2 and GPT-4 exhibit exceptional capabilities in text capture, understanding, and natural language generation, thanks to deep learning algorithms trained on vast amounts of textual data\cite{huang_survey_2024,touvron_llama_nodate,openai_gpt-4_2024}. These models efficiently perform natural language processing tasks such as text summarization, question answering, and translation. By leveraging these models, we can achieve more efficient and accurate understanding and labeling of animal behavior through natural language interaction, providing valuable theoretical support and practical assistance in promoting animal welfare.

However, given the multimodal nature of information in real-world environments, LLMs that accept only textual user inputs and responses are insufficient to meet the diverse needs of application scenarios\cite{huang_language_2023,zhang_speechgpt_2023}. With the continuous advancement of artificial intelligence technology, multimodal large models based on prompts have introduced new opportunities for development in natural language understanding and generation\cite{yin_survey_2024}. By incorporating pre-training on image-text and audio-text data, these models have achieved comprehensive processing capabilities for video, audio, and textual data, significantly enhancing the understanding of multimodal content by LLMs\cite{sun_emu_2024,li_blip-2_2023}. Recent research has begun to explore the capabilities of multimodal LLMs in processing dynamic visual content and textual inputs to promote a deep understanding of video content\cite{li_videochat_2024,maaz_video-chatgpt_2023}. Models such as Video-LLaMA, Video-Chat2, and the latest GPT-4 omni (GPT-4o) have demonstrated cross-modal reasoning abilities spanning audio, vision, and text, injecting new vitality into the field of natural language processing\cite{zhang_video-llama_2023,li_mvbench_2024,noauthor_hello_nodate}.

Visual perception is an essential part of human cognition, encompassing the perception of object shapes, colors, depths, and other features, as well as advanced processing capabilities such as object recognition, comparison, description, and memory\cite{marr_vision_1983}. By capturing and analyzing light signals in the environment, humans can quickly identify objects, judge spatial positions, track dynamic changes, and memorize and compare similar objects. Simulating and enhancing this visual perception ability has become a research focus in computer vision. Tasks such as image recognition, object detection, and image segmentation rely on vast amounts of data and advanced algorithms to mimic human visual processing mechanisms\cite{lowe_object_1999,wang_layered_1993}. These techniques aim to enable computers to understand and interpret visual information like humans, thus playing a role in various application scenarios. However, visual perception, as a critical component for multimodal models to complete tasks, also faces significant challenges\cite{luo_feast_2024,pan_auto-encoding_2024,yue_mmmu_2023,liu_mmbench_2024}. Since visual information cannot be directly mediated through natural language, multimodal large models struggle to integrate and analyze data from different modalities\cite{berrios_towards_2023}. Therefore, effectively fusing visual information with other modal details and enhancing the processing capabilities of multimodal models is a critical aspect of current multimodal research.

The livestock industry is a crucial area where computer vision technologies, particularly LLMs, demonstrate significant application potential. These technologies can bring unprecedented efficiency, intelligent management, and decision support to the livestock industry\cite{li_barriers_2022}. However, accurately identifying changes in animal behavior in complex farming environments remains a significant challenge for further advancements in innovative farming technology. Given the multimodal nature of the work, from traditional animal behavior semantic definitions to video behavior understanding, combining human visual understanding characteristics with state-of-the-art (SOTA) multimodal LLMs may provide a new direction for rapid development in animal behavior understanding. This study explores the application potential and limitations of current multimodal large models in animal behavior understanding and innovative farming by comparing and analyzing the visual perception capabilities of these models for video clips of different piglet activity types. 

\section{Related work}

The rapid development of natural language processing and LLMs has provided new opportunities for cross-domain integration. Specifically, integrating visual encoders into pre-trained LLMs has led to significant advancements in visual-language LLMs\cite{li_blip-2_2023}. A series of studies have focused on applying multimodal LLMs to explore the possibilities of video understanding further. As a sequence of frames with a temporal dimension, the complexity and challenges of video often render single-frame analysis insufficient to provide satisfactory solutions. Therefore, multimodal LLMs must understand visual and textual information jointly and fully utilize dynamic temporal information in video data\cite{liu_visual_2023,guan_hallusionbench_2024}. This capability can extend the success of LLMs in image understanding to more complex video scenarios.

To enhance the visual and auditory capabilities of large LLMs in video understanding, DAMO Academy (Alibaba Group) proposed Video-LLaMA, an innovative audio-visual language model\cite{zhang_video-llama_2023}. This multimodal framework effectively maps video frames and audio signals into the textual input of LLMs, enabling cross-modal instruction. Through this approach, Video-LLaMA has completed complex tasks such as audio/video description, writing, and question answering, demonstrating its superior performance and broad application prospects. Unlike previous large models focused on static image understanding, such as Mini-GPT4 and LLaVA\cite{liu_visual_2023,zhu_minigpt-4_2023}, Video-LLaMA can capture dynamic scene changes in videos and integrate audio-visual signals. This capability enables Video-LLaMA to process video content more accurately and efficiently, bringing breakthroughs to video processing and analysis.

Meanwhile, MiniGPT4-Video, proposed by research teams from KAUST and Harvard University, demonstrates strong video understanding potential\cite{ataallah_minigpt4-video_2024}. This model can simultaneously process temporal visual and textual data, extending the ability of MiniGPT-v2 to convert visual features from single images into the LLM space\cite{chen_minigpt-v2_2023}, enabling it to understand video content. Through large-scale image-text paired pre-training, MiniGPT4-Video maps the features encoded by the visual encoder to the textual space of the language model. It utilizes multi-frame video inputs for video-text paired pre-training, achieving an in-depth understanding and modeling of video content.

Video-Chat2, one of the current SOTA video-multimodal LLMs, has effectively improved the effectiveness of dynamic task processing through enhancements in temporal perception and cognition\cite{li_mvbench_2024}. The model incorporates a wide range of image and video data into its instruction set, increasing the diversity of instruction adjustment data for progressive multimodal training. Like Video-LLaMA, it extends the BLIP-2 architecture for video embedding extraction, adopting a dual-stream approach that combines audio and visual signals with visual and semantic information. Video-Chat2 can generate natural and relevant dialogues based on the understanding of video content, providing a more fluid and natural experience for human-computer interaction. With the continuous advancement of artificial intelligence technology, natural language processing (NLP) has also developed significantly. As the successor to the GPT series, the latest GPT-4 omni (GPT-4o) has further optimized and expanded its capabilities, enhancing the model’s understanding and generation abilities\cite{noauthor_hello_nodate}. GPT-4o can accept any combination of text, audio, and image inputs and outputs to handle more complex language tasks. This versatile input-output capability makes GPT-4o more similar to natural human-to-human communication in human-computer interaction, injecting new vitality into the development of artificial intelligence technology.

\section{Material and method}

\subsection{Sample Data}
The animal video data used in this study were captured in July 2021. A camera was installed above the pen to capture the activities of 16 pigs from the top view, with a video resolution of 3840 × 2160 pixels and a frame rate of 15 FPS. To protect the privacy and focus on the analysis, all videos in this study were masked to remove human identification and other irrelevant visual elements, and the resolution was uniformly adjusted to 1080 × 720 pixels. Video data were collected continuously. The lighting conditions in the house remained constant to mitigate the impact of light changes on video quality. An equal-interval fixed-point sampling technique was employed to prepare the sample data to preserve the time-series data's integrity and increase the data processing effectiveness. A representative sample was taken from every 2nd minute of video data over a continuous 24-hour period, with a sampling interval of 10 minutes, resulting in a total of 144 one-minute video samples.

To achieve individual animal tracking, this experiment employed the Decoupled Video Segmentation Approach (DEVA)\cite{cheng_tracking_2023}. This method combines image-level segmentation with class/task-agnostic bi-directional temporal propagation characteristics, effectively applying it to track individual pigs in 1-minute video samples. Specifically, DEVA segments each frame of the video, identifying and extracting the contours and features of individual pigs. Simultaneously, through bi-directional temporal propagation technology, DEVA can span the time intervals between different frames, maintaining continuous tracking of the same individual pig. The video was divided into segments based on the DEVA algorithm’s pig tracking results. Each segment included a whole behavior cycle, guaranteeing that the beginning and finish of the target behavior were contained inside it. Ultimately, 18 video clips depicting various behaviors, such as standing, feeding, drinking, lying, moving, and socializing, were chosen to preliminarily test the visual perception of multimodel LLMs’ ability, with three video clips representing each category of behavior.

\subsection{Visual Prompt}
Current research primarily focuses on the text reasoning capabilities of multimodal LLMs, while visual cues also play a pivotal role in shaping their performance\cite{fu_blink_2024}. To investigate the impact of visual cues on the understanding of piglet activities and the visual perception capabilities of multimodal LLMs, two types of test data based on 18 video clips were created: close-up video clips of individual pigs and full-shot video clips with marks (Appendix I). The close-up video clips mostly show the target animal’s whole activity. By eliminating much of the backdrop, they highlight the actor’s behavior and act as a visual signal. Several animal individuals, including the target individual, are shown in the marked full-shot video clips. Marks on the target individual and important elements in the scenario serve as additional visual cues. In this investigation, the ability of multimodal LLM models to comprehend videos in which the behavior actor is highlighted, and all of the scene information is provided is assessed using these two kinds of video clips.

This study designed two prompt templates corresponding to two types of video clips to investigate the comprehension abilities of multimodal LLMs for piglet activities, reveal their enormous potential in natural language processing and video understanding, and offer fresh insights and solutions for applications in animal husbandry (Appendix II). The features of the livestock scene data and the ensuing application needs serve as the foundation for these templates. As a common way of prompting in LLMs, a pure text prompt template is used for the close-up video clips of individual pigs. All of the descriptions in this template are textual; they provide the necessary context, describe the situation, and then ask specific questions regarding the animal’s behavior in the video. General text and image marks are required for the full-shot video clips with marks to draw attention to the behavior actor and important spots associated with piglet activities. To help the model focus on the target animal and particular behavioral actions, the prompt template highlights these visual cues by pointing out specific areas in the video. In the full-shot video clips of this experiment, the activity actor (the target pig) was marked with a red mask. After marking the drinking and feeding areas in the frame with green and blue polygons, subsequent questions were asked.

\subsection{Evaluation Metrics}

As indicated in Table~\ref{em}, this study proposed five evaluation metrics to thoroughly assess various SOTA multimodal LLMs’ visual perception and video comprehension skills in livestock scenarios. The evaluation metrics, grounded in human understanding, offer relative scores for the models’ predictions from 0 to 5, signifying the level of agreement between the model’s output and the actual data. Specifically, each video clip’s prediction result were evaluated as either 0, 0.5 or 1, adhering strictly to the predefined rules. Subsequently, to standardize the evaluation metrics, the outcomes of the 18 video clips were normalized within a range of 0 to 5. A score of 5 denotes near consistency, whereas a score of 0 indicates a substantial departure from the ground truth. Identical prompt templates and accompanying test data were used throughout the evaluation to accurately compare the outcomes from four multimodal LLMs.

\begin{table}[h]
    \centering
    \renewcommand{\arraystretch}{1.25}
     \caption{Evaluation metrics and description.}
    \begin{tabular}{c l}
    \hline %
         \textbf{Evaluation Metrics}&\textbf{Description}  \\
    \hline
         Counting (Full-shot video clips)&i. Ratio of correctly detected animals \\
         Actor referring	&i. Proportion of correctly identified actors \\
         &i. Accuracy of behavior classification \\
Semantic correspondence&ii. Presence of relevant keywords\\
&iii. Descriptive text length\\
Time perception	&i. Duration of the behavioral occurrence\\
Robustness	&i. Consistency of results among three identical behavior types \\  &(from different videos)\\
\hline
    \end{tabular}
    \label{em}
\end{table}

Livestock scenarios are complex, with overlapping animals, occlusions, and differences in size and appearance. In terms of detection, identification, and compositional reasoning, multimodal LLMs should be primarily capable of accurately identifying the number of animals in video clips and the target animal in the scene. As a result, the accuracy of counting and actor referring were used in this paper’s evaluation metrics. Additionally, semantic correspondence was used to measure the model’s performance in animal behavior classification, text description, and the inclusion of behavior-related keywords. Time perception assesses the model’s ability to integrate temporal information about the target content in video clips while answering questions. Robustness evaluates the consistency of the model’s results when processing different but similar video clips, reflecting its stability and reliability.

\begin{figure}[h]
\includegraphics[width=14cm]{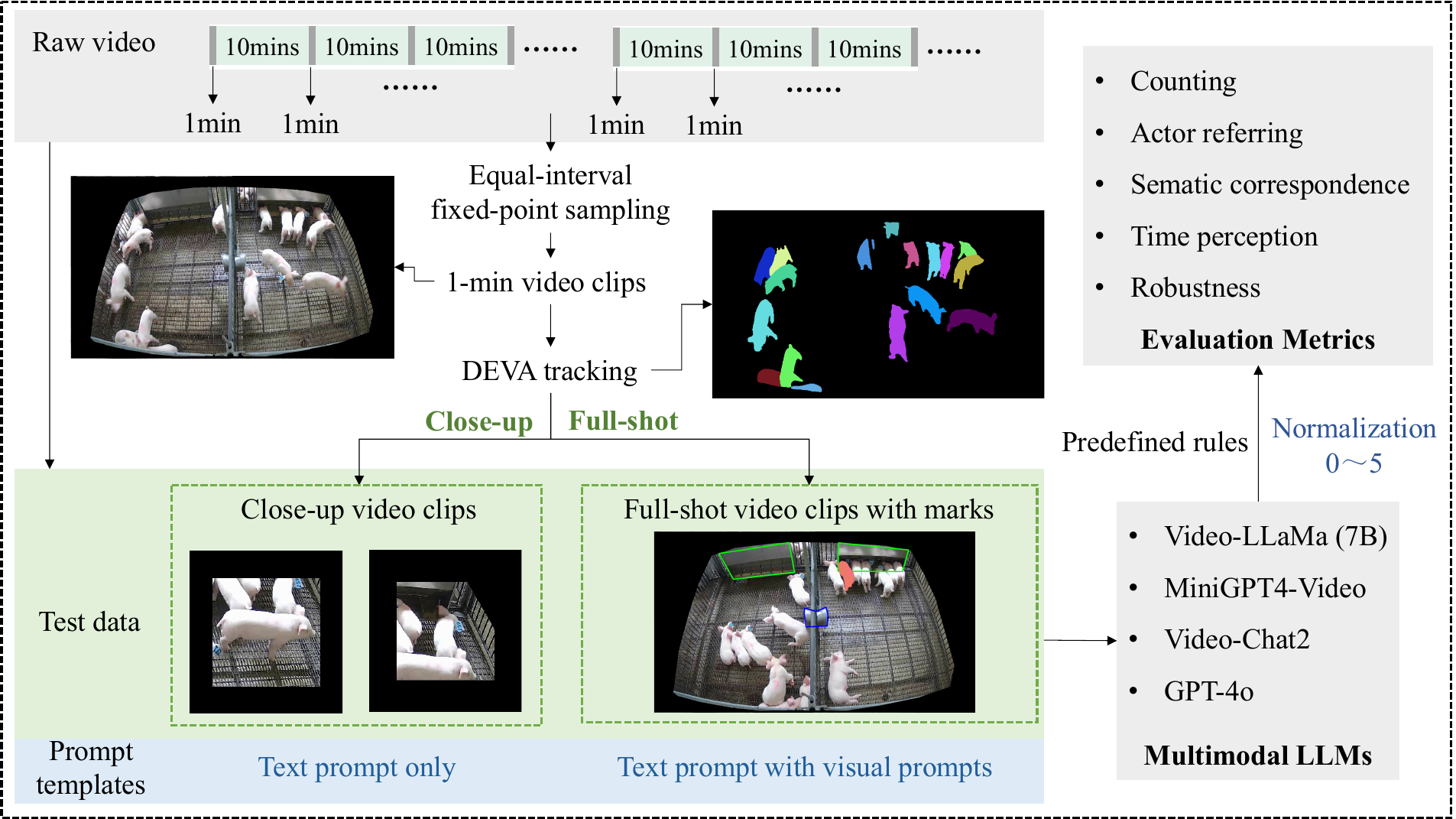}
\caption{Experiment workflow.}
\label{workflow}
\end{figure}

In this study, two types of prompt templates mentioned above were evaluated, along with the corresponding video data, on four multimodal LLMs: Video-LLaMA (7B-version), MiniGPT4-Video, Video-Chat2, and GPT-4o. The hyperparameter temperature was uniformly set to 1 across all models for consistency. Afterward, predefined rules were employed to extract the outcomes for the evaluation metrics from the model outputs. A quantization approach was implemented to improve the comparability and interpretability of the prediction results from the four multimodal LLMs that were being evaluated. The scores were adjusted to fall between 0 and 5. Figure~\ref{workflow} shows the workflow for this experiment.

\section{Results and discussions}

\begin{figure}[h]
\includegraphics[width=14cm]{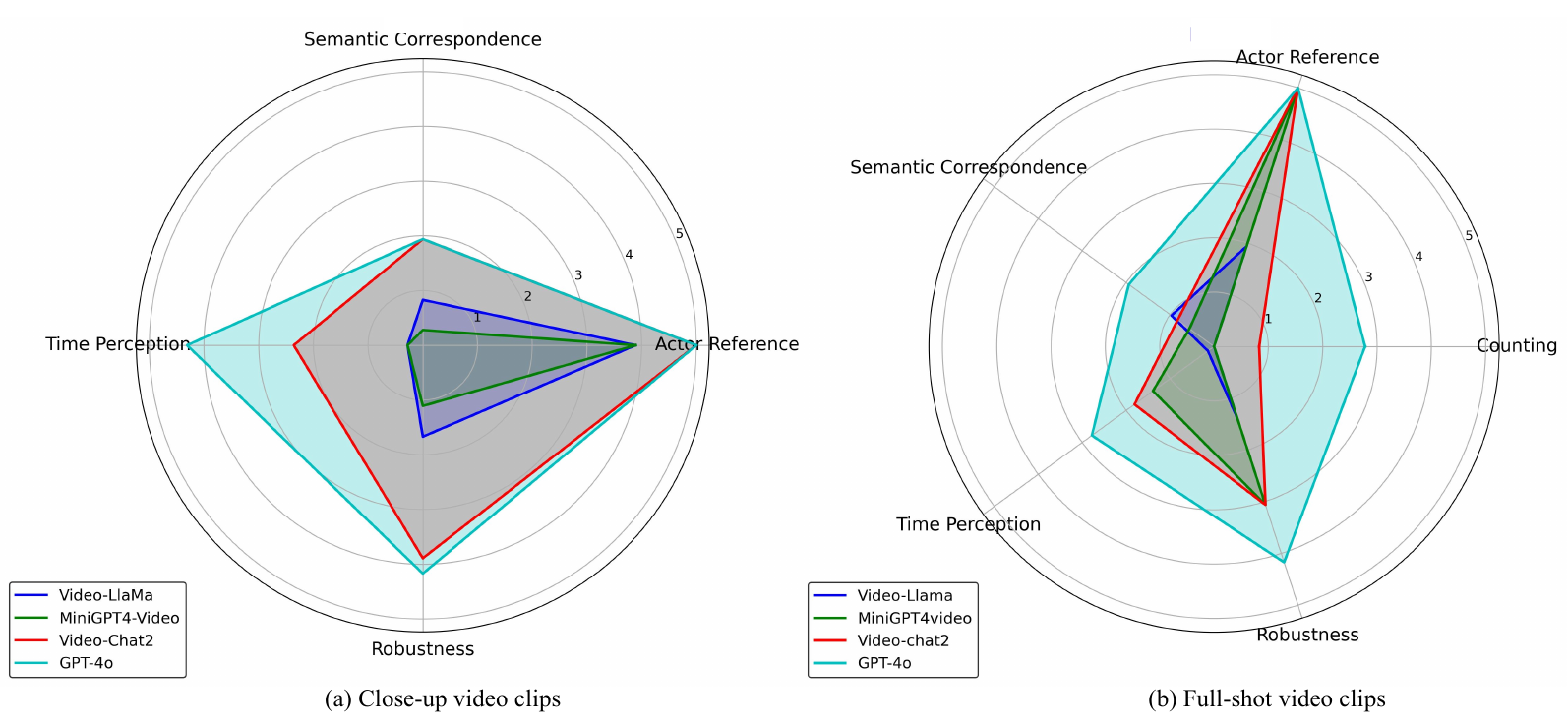}
\caption{Evaluation results of multimodal LLMs on two types of video clips.}
\label{result}
\end{figure}

The evaluation results of the four multimodal LLMs on close-up video clips and full-shot video clips are presented in Figure~\ref{result}. In understanding piglet activity in livestock scenarios, the four models tested in this experiment demonstrated a certain degree of actor referring ability in both close-up and full-shot video clips, enabling them to identify and locate the target piglets. Moreover, these models showed potential in terms of robustness and time perception. However, all four models performed poorly in semantic correspondence, which is most closely associated with the ability of multimodal LLMs to interpret animal activities in videos among the five evaluation metrics. The detailed scoring results for the five aspects of the evaluation metrics are shown in Table~\ref{score}. 

\begin{table}[h]
    \centering
    \renewcommand{\arraystretch}{1.25}
     \caption{Scoring results of four multimodal LLMs.}
    \begin{tabular}{ccccccccc}
    \hline %
      & \multicolumn{8}{c}{Multimodal LLMs} \\ 
    \textbf{Evaluation Metrics} &\multicolumn{2}{|c}{\textbf{Video-Llama}}&\multicolumn{2}{c}{\textbf{MiniGPT4video}} &\multicolumn{2}{c}{\textbf{Video-chat2}}&\multicolumn{2}{c}{\textbf{GPT-4o}}\\
&\textit{C}\footnotemark[1]&\textit{F}\footnotemark[2]&\textit{C}&\textit{F}&\textit{C}&\textit{F}&\textit{C}&\textit{F}\\
\hline
Counting& -&	0&	-&	0&	-&	0.83&	-&	2.78\\
Actor Referring&	3.89&	1.94&	3.89&	5&	5&	5&	5&	5\\
Semantic Correspondence&	0.83&	0.97&	0.28&	0.56&	\textbf{1.94}&	0.83&	\textbf{1.94}&	1.94\\
Time Perception&	0.28&	0.14&	0.28&	1.39&	\textbf{2.36}&	1.81&	\textbf{4.31}&	2.78\\
Robustness&	1.67&	1.39&	1.11&	3.06&	3.89&	3.06&	4.17&	4.17\\
\hline
    \end{tabular}
    \label{score}
     {\small {\textit{[1]C:Close-up video clips.}}{\textit{[2]F:Full-shot video clips.}}} 
\end{table}

\paragraph{Close-up video clips:} In evaluating the four multimodal LLMs on close-up video clips, Video-LLaMA and MiniGPT4-Video exhibited similar overall performance. However, both models displayed low sensitivity in time perception, scoring 0.28, indicating difficulties in identifying the duration of pig activities or providing accurate duration estimates. In contrast, Video-Chat2 and GPT-4o demonstrated superior performance, particularly regarding time perception and robustness. Specifically, all four multimodal LLMs performed relatively well in actor referring, accurately identifying the actor of the activity. However, these four models generally performed poorly in semantic correspondence, with evaluation scores less than 2. Additionally, Video-Chat2 and GPT-4o significantly outperformed Video-LLaMA and MiniGPT4-Video in time perception and robustness, with Video-Chat2 scoring 2.36 and 3.89, and GPT-4o scoring 4.31 and 4.17, respectively.

\paragraph{Full-shot video clips:}In evaluating the comprehension abilities of the four multimodal LLMs on full-shot video clips, counting was introduced as an additional evaluation metric. Based on the comprehensive evaluation results, MiniGPT4-Video and Video-Chat2 exhibited similar performance across multiple evaluation aspects, while Video-LLaMA performed relatively poorer. Among the four models, GPT-4o achieved significant advantages in all five evaluation dimensions, particularly in counting, where its score surpassed Video-Chat2 by 1.95 points. Notably, both Video-LLaMA and MiniGPT4-Video scored 0 on the counting metric. Further analysis revealed that MiniGPT4-Video, Video-Chat2, and GPT-4o demonstrated a certain level of recognition and understanding in actor referring, time perception, and robustness, all performing better than Video-LLaMA. However, in terms of semantic correspondence, all four models exhibited deficiencies, with only GPT-4o scoring close to 2, indicating a relative advantage compared to the other models.

\begin{figure}[h]
    \centering
    \includegraphics[width=14cm]{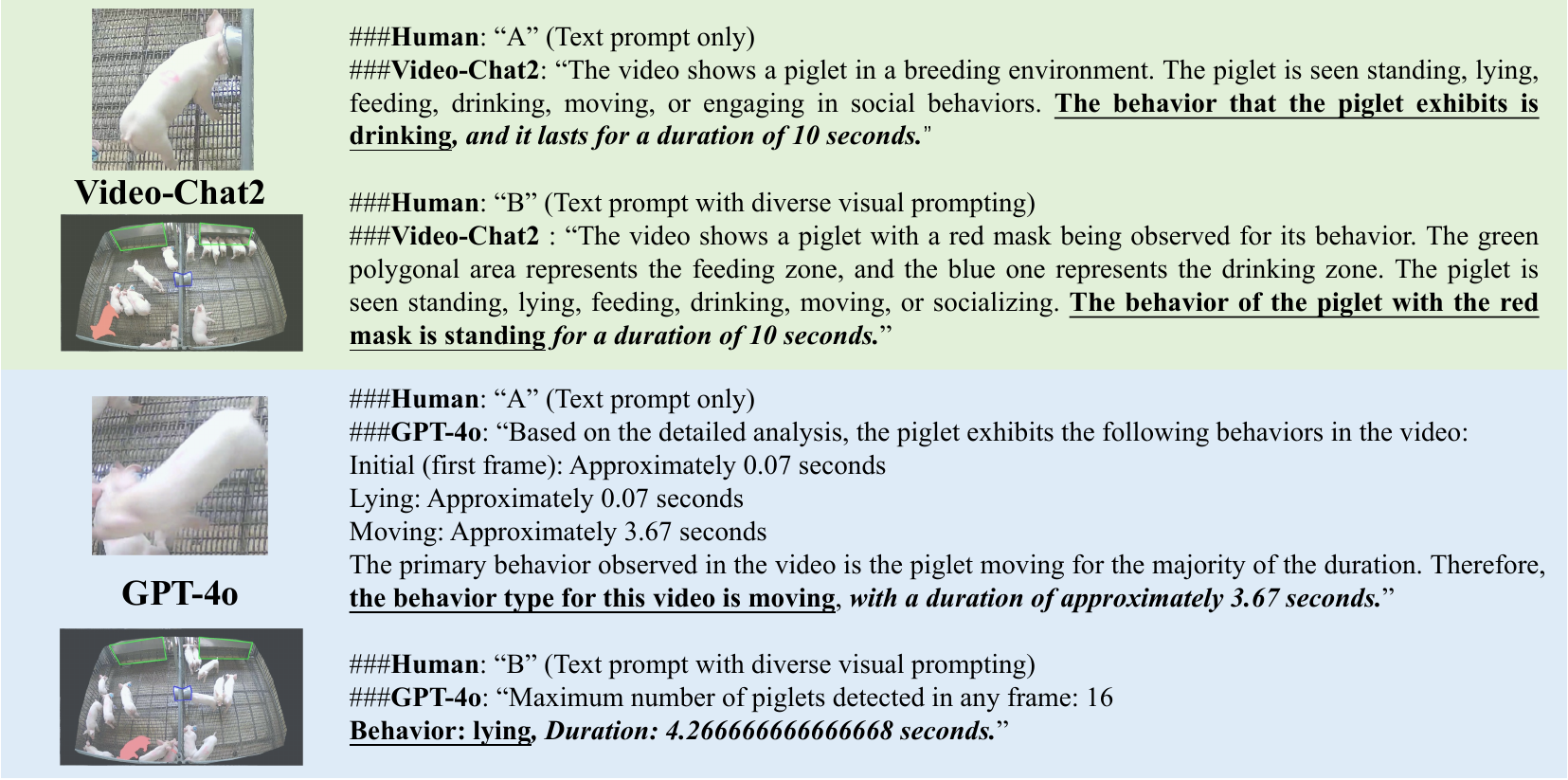}
    \caption{Examples of Video-Chat2 and GPT-4o.}
    \label{exp}
\end{figure}

Notably, Video-Chat2 and GPT-4o demonstrated superior performance in semantic correspondence and time perception. As illustrated in Figure~\ref{exp}, these two models directly output the type of behavior (as highlighted by bold font with an underline), and the responses were concise. The average number of words in their answers were 287 and 246, respectively, compared to 735 and 337 for Video-LLaMA and MiniGPT4-Video. Despite Video-Chat2 and GPT-4o having an understanding of temporal tasks (as highlighted by italics), these two models showed limitations in the accuracy of time perception. This discrepancy indicates that their temporal understanding modules require further refinement.

In specific tasks such as understanding piglet activity, humans can accurately analyze animal behaviors in complex farming scenarios, yet this remains challenging for current multimodal LLM models. Despite these models having a solid foundation in video understanding, question answering, and generating rich contexts, their performance in specialized applications like animal activity understanding still needs to improve, even for advanced models like GPT-4o. This indicates that further optimization and enhancement are necessary for existing multimodal LLMs to handle more complex and specialized application scenarios effectively.

\section{Conclusion}
This study tested two types of test data: close-up and full-shot video clips, and corresponding prompt templates were designed for these video clips. Based on this, a comprehensive evaluation of the visual perception abilities of four multimodal LLMs was conducted: Video-LLaMA (7B version), MiniGPT4-Video, Video-Chat2, and GPT-4o, in the task of piglet activity understanding. The evaluation results demonstrated that multimodal LLMs exhibit significant visual perception potential in animal activity video understanding, with GPT-4o and Video-Chat2 models standing out relatively. Notably, the close-up video clips used in this experiment proved more suitable for video understanding tasks of multimodal LLMs, providing a new perspective and effective tool for animal activity understanding in livestock scenarios.

However, despite the potential demonstrated by these multimodal LLMs in piglet activity understanding, they still need to improve to become specialized models regarding livestock scene understanding. Specifically, the current models require further enhancement in animal activity perception capabilities and occasionally exhibit ‘hallucinations’ when processing animal activities in livestock scenes, where the predicted results do not align with the actual animal activities or scenes. In the future, as research on multimodal LLMs progresses and their functions are continuously improved, we expect these models to achieve more significant breakthroughs in specialized tasks such as animal behavior understanding. To this end, we will continue to explore optimization strategies for visual prompts and prompt templates to adapt to the complexity and diversity of livestock scenes, laying a solid foundation for applying multimodal LLMs in livestock scene understanding.

\medskip
\newpage
{\small

\begin{thebibliography}{10}

\bibitem{ataallah_minigpt4-video_2024}
K.~Ataallah, X.~Shen, E.~Abdelrahman, E.~Sleiman, D.~Zhu, J.~Ding, and M.~Elhoseiny, ``{MiniGPT4}-{Video}: {Advancing} {Multimodal} {LLMs} for {Video} {Understanding} with {Interleaved} {Visual}-{Textual} {Tokens},'' Apr. 2024.

\bibitem{fu_blink_2024}
X.~Fu, Y.~Hu, B.~Li, Y.~Feng, H.~Wang, X.~Lin, D.~Roth, N.~A. Smith, W.-C. Ma, and R.~Krishna, ``{BLINK}: {Multimodal} {Large} {Language} {Models} {Can} {See} but {Not} {Perceive},'' May 2024.

\bibitem{zhang_video-llama_2023}
H.~Zhang, X.~Li, and L.~Bing, ``Video-{LLaMA}: {An} {Instruction}-tuned {Audio}-{Visual} {Language} {Model} for {Video} {Understanding},'' Oct. 2023.

\bibitem{huang_survey_2024}
K.~Huang, F.~Mo, H.~Li, Y.~Li, Y.~Zhang, W.~Yi, Y.~Mao, J.~Liu, Y.~Xu, J.~Xu, J.-Y. Nie, and Y.~Liu, ``A {Survey} on {Large} {Language} {Models} with {Multilingualism}: {Recent} {Advances} and {New} {Frontiers},'' May 2024.

\bibitem{li_videochat_2024}
K.~Li, Y.~He, Y.~Wang, Y.~Li, W.~Wang, P.~Luo, Y.~Wang, L.~Wang, and Y.~Qiao, ``{VideoChat}: {Chat}-{Centric} {Video} {Understanding},'' Jan. 2024.

\bibitem{chen_minigpt-v2_2023}
J.~Chen, D.~Zhu, X.~Shen, X.~Li, Z.~Liu, P.~Zhang, R.~Krishnamoorthi, V.~Chandra, Y.~Xiong, and M.~Elhoseiny, ``{MiniGPT}-v2: large language model as a unified interface for vision-language multi-task learning,'' Oct. 2023.

\bibitem{li_mvbench_2024}
K.~Li, Y.~Wang, Y.~He, Y.~Li, Y.~Wang, Y.~Liu, Z.~Wang, J.~Xu, G.~Chen, P.~Luo, L.~Wang, and Y.~Qiao, ``{MVBench}: {A} {Comprehensive} {Multi}-modal {Video} {Understanding} {Benchmark},'' Jan. 2024.

\bibitem{maaz_video-chatgpt_2023}
M.~Maaz, H.~Rasheed, S.~Khan, and F.~S. Khan, ``Video-{ChatGPT}: {Towards} {Detailed} {Video} {Understanding} via {Large} {Vision} and {Language} {Models},'' June 2023.

\bibitem{openai_gpt-4_2024}
{OpenAI}, J.~Achiam, S.~Adler, S.~Agarwal, L.~Ahmad, I.~Akkaya, F.~L. Aleman, D.~Almeida, J.~Altenschmidt, S.~Altman, S.~Anadkat, R.~Avila, I.~Babuschkin, S.~Balaji, V.~Balcom, P.~Baltescu, H.~Bao, M.~Bavarian, J.~Belgum, I.~Bello, J.~Berdine, G.~Bernadett-Shapiro, C.~Berner, L.~Bogdonoff, O.~Boiko, M.~Boyd, A.-L. Brakman, G.~Brockman, T.~Brooks, M.~Brundage, K.~Button, T.~Cai, R.~Campbell, A.~Cann, B.~Carey, C.~Carlson, R.~Carmichael, B.~Chan, C.~Chang, F.~Chantzis, D.~Chen, S.~Chen, R.~Chen, J.~Chen, M.~Chen, B.~Chess, C.~Cho, C.~Chu, H.~W. Chung, D.~Cummings, J.~Currier, Y.~Dai, C.~Decareaux, T.~Degry, N.~Deutsch, D.~Deville, A.~Dhar, D.~Dohan, S.~Dowling, S.~Dunning, A.~Ecoffet, A.~Eleti, T.~Eloundou, D.~Farhi, L.~Fedus, N.~Felix, S.~P. Fishman, J.~Forte, I.~Fulford, L.~Gao, E.~Georges, C.~Gibson, V.~Goel, T.~Gogineni, G.~Goh, R.~Gontijo-Lopes, J.~Gordon, M.~Grafstein, S.~Gray, R.~Greene, J.~Gross, S.~S. Gu, Y.~Guo, C.~Hallacy, J.~Han, J.~Harris, Y.~He, M.~Heaton, J.~Heidecke, C.~Hesse, A.~Hickey,
  W.~Hickey, P.~Hoeschele, B.~Houghton, K.~Hsu, S.~Hu, X.~Hu, J.~Huizinga, S.~Jain, S.~Jain, J.~Jang, A.~Jiang, R.~Jiang, H.~Jin, D.~Jin, S.~Jomoto, B.~Jonn, H.~Jun, T.~Kaftan, Å.~Kaiser, A.~Kamali, I.~Kanitscheider, N.~S. Keskar, T.~Khan, L.~Kilpatrick, J.~W. Kim, C.~Kim, Y.~Kim, J.~H. Kirchner, J.~Kiros, M.~Knight, D.~Kokotajlo, Å.~Kondraciuk, A.~Kondrich, A.~Konstantinidis, K.~Kosic, G.~Krueger, V.~Kuo, M.~Lampe, I.~Lan, T.~Lee, J.~Leike, J.~Leung, D.~Levy, C.~M. Li, R.~Lim, M.~Lin, S.~Lin, M.~Litwin, T.~Lopez, R.~Lowe, P.~Lue, A.~Makanju, K.~Malfacini, S.~Manning, T.~Markov, Y.~Markovski, B.~Martin, K.~Mayer, A.~Mayne, B.~McGrew, S.~M. McKinney, C.~McLeavey, P.~McMillan, J.~McNeil, D.~Medina, A.~Mehta, J.~Menick, L.~Metz, A.~Mishchenko, P.~Mishkin, V.~Monaco, E.~Morikawa, D.~Mossing, T.~Mu, M.~Murati, O.~Murk, D.~Mély, A.~Nair, R.~Nakano, R.~Nayak, A.~Neelakantan, R.~Ngo, H.~Noh, L.~Ouyang, C.~O'Keefe, J.~Pachocki, A.~Paino, J.~Palermo, A.~Pantuliano, G.~Parascandolo, J.~Parish, E.~Parparita, A.~Passos,
  M.~Pavlov, A.~Peng, A.~Perelman, F.~d. A.~B. Peres, M.~Petrov, H.~P. d.~O. Pinto, {Michael}, {Pokorny}, M.~Pokrass, V.~H. Pong, T.~Powell, A.~Power, B.~Power, E.~Proehl, R.~Puri, A.~Radford, J.~Rae, A.~Ramesh, C.~Raymond, F.~Real, K.~Rimbach, C.~Ross, B.~Rotsted, H.~Roussez, N.~Ryder, M.~Saltarelli, T.~Sanders, S.~Santurkar, G.~Sastry, H.~Schmidt, D.~Schnurr, J.~Schulman, D.~Selsam, K.~Sheppard, T.~Sherbakov, J.~Shieh, S.~Shoker, P.~Shyam, S.~Sidor, E.~Sigler, M.~Simens, J.~Sitkin, K.~Slama, I.~Sohl, B.~Sokolowsky, Y.~Song, N.~Staudacher, F.~P. Such, N.~Summers, I.~Sutskever, J.~Tang, N.~Tezak, M.~B. Thompson, P.~Tillet, A.~Tootoonchian, E.~Tseng, P.~Tuggle, N.~Turley, J.~Tworek, J.~F.~C. Uribe, A.~Vallone, A.~Vijayvergiya, C.~Voss, C.~Wainwright, J.~J. Wang, A.~Wang, B.~Wang, J.~Ward, J.~Wei, C.~J. Weinmann, A.~Welihinda, P.~Welinder, J.~Weng, L.~Weng, M.~Wiethoff, D.~Willner, C.~Winter, S.~Wolrich, H.~Wong, L.~Workman, S.~Wu, J.~Wu, M.~Wu, K.~Xiao, T.~Xu, S.~Yoo, K.~Yu, Q.~Yuan, W.~Zaremba, R.~Zellers,
  C.~Zhang, M.~Zhang, S.~Zhao, T.~Zheng, J.~Zhuang, W.~Zhuk, and B.~Zoph, ``{GPT}-4 {Technical} {Report},'' Mar. 2024.

\bibitem{touvron_llama_nodate}
H.~Touvron, L.~Martin, and K.~Stone, ``Llama 2: {Open} {Foundation} and {Fine}-{Tuned} {Chat} {Models},''

\bibitem{noauthor_hello_nodate}
``Hello {GPT}-4o.''

\bibitem{huang_language_2023}
S.~Huang, L.~Dong, W.~Wang, Y.~Hao, S.~Singhal, S.~Ma, T.~Lv, L.~Cui, O.~K. Mohammed, B.~Patra, Q.~Liu, K.~Aggarwal, Z.~Chi, J.~Bjorck, V.~Chaudhary, S.~Som, X.~Song, and F.~Wei, ``Language {Is} {Not} {All} {You} {Need}: {Aligning} {Perception} with {Language} {Models},'' Mar. 2023.

\bibitem{zhang_speechgpt_2023}
D.~Zhang, S.~Li, X.~Zhang, J.~Zhan, P.~Wang, Y.~Zhou, and X.~Qiu, ``{SpeechGPT}: {Empowering} {Large} {Language} {Models} with {Intrinsic} {Cross}-{Modal} {Conversational} {Abilities},'' May 2023.

\bibitem{yin_survey_2024}
S.~Yin, C.~Fu, S.~Zhao, K.~Li, X.~Sun, T.~Xu, and E.~Chen, ``A {Survey} on {Multimodal} {Large} {Language} {Models},'' Apr. 2024.

\bibitem{sun_emu_2024}
Q.~Sun, Q.~Yu, Y.~Cui, F.~Zhang, X.~Zhang, Y.~Wang, H.~Gao, J.~Liu, T.~Huang, and X.~Wang, ``Emu: {Generative} {Pretraining} in {Multimodality},'' May 2024.

\bibitem{li_blip-2_2023}
J.~Li, D.~Li, S.~Savarese, and S.~Hoi, ``{BLIP}-2: {Bootstrapping} {Language}-{Image} {Pre}-training with {Frozen} {Image} {Encoders} and {Large} {Language} {Models},'' June 2023.

\bibitem{marr_vision_1983}
D.~Marr, ``Vision: {A} {Computational} {Investigation} into the {Human} {Representation} and {Processing} of {Visual} {Information},'' 1983.

\bibitem{lowe_object_1999}
D.~Lowe, ``Object recognition from local scale-invariant features,'' in {\em Proceedings of the {Seventh} {IEEE} {International} {Conference} on {Computer} {Vision}}, vol.~2, pp.~1150--1157 vol.2, Sept. 1999.

\bibitem{wang_layered_1993}
J.~Wang and E.~Adelson, ``Layered representation for motion analysis,'' in {\em Proceedings of {IEEE} {Conference} on {Computer} {Vision} and {Pattern} {Recognition}}, (New York, NY, USA), pp.~361--366, IEEE Comput. Soc. Press, 1993.

\bibitem{luo_feast_2024}
G.~Luo, Y.~Zhou, Y.~Zhang, X.~Zheng, X.~Sun, and R.~Ji, ``Feast {Your} {Eyes}: {Mixture}-of-{Resolution} {Adaptation} for {Multimodal} {Large} {Language} {Models},'' Mar. 2024.

\bibitem{pan_auto-encoding_2024}
K.~Pan, S.~Tang, J.~Li, Z.~Fan, W.~Chow, S.~Yan, T.-S. Chua, Y.~Zhuang, and H.~Zhang, ``Auto-{Encoding} {Morph}-{Tokens} for {Multimodal} {LLM},'' May 2024.

\bibitem{yue_mmmu_2023}
X.~Yue, Y.~Ni, K.~Zhang, T.~Zheng, R.~Liu, G.~Zhang, S.~Stevens, D.~Jiang, W.~Ren, Y.~Sun, C.~Wei, B.~Yu, R.~Yuan, R.~Sun, M.~Yin, B.~Zheng, Z.~Yang, Y.~Liu, W.~Huang, H.~Sun, Y.~Su, and W.~Chen, ``{MMMU}: {A} {Massive} {Multi}-discipline {Multimodal} {Understanding} and {Reasoning} {Benchmark} for {Expert} {AGI},'' Dec. 2023.

\bibitem{liu_mmbench_2024}
Y.~Liu, H.~Duan, Y.~Zhang, B.~Li, S.~Zhang, W.~Zhao, Y.~Yuan, J.~Wang, C.~He, Z.~Liu, K.~Chen, and D.~Lin, ``{MMBench}: {Is} {Your} {Multi}-modal {Model} an {All}-around {Player}?,'' Apr. 2024.

\bibitem{berrios_towards_2023}
W.~Berrios, G.~Mittal, T.~Thrush, D.~Kiela, and A.~Singh, ``Towards {Language} {Models} {That} {Can} {See}: {Computer} {Vision} {Through} the {LENS} of {Natural} {Language},'' June 2023.

\bibitem{liu_visual_2023}
H.~Liu, C.~Li, Q.~Wu, and Y.~J. Lee, ``Visual {Instruction} {Tuning},'' Dec. 2023.

\bibitem{guan_hallusionbench_2024}
T.~Guan, F.~Liu, X.~Wu, R.~Xian, Z.~Li, X.~Liu, X.~Wang, L.~Chen, F.~Huang, Y.~Yacoob, D.~Manocha, and T.~Zhou, ``{HallusionBench}: {An} {Advanced} {Diagnostic} {Suite} for {Entangled} {Language} {Hallucination} and {Visual} {Illusion} in {Large} {Vision}-{Language} {Models},'' Mar. 2024.

\bibitem{zhu_minigpt-4_2023}
D.~Zhu, J.~Chen, X.~Shen, X.~Li, and M.~Elhoseiny, ``{MiniGPT}-4: {Enhancing} {Vision}-{Language} {Understanding} with {Advanced} {Large} {Language} {Models},'' Oct. 2023.

\bibitem{cheng_tracking_2023}
H.~K. Cheng, S.~W. Oh, B.~Price, A.~Schwing, and J.-Y. Lee, ``Tracking {Anything} with {Decoupled} {Video} {Segmentation},'' Sept. 2023.

\bibitem{li_barriers_2022}
J.~Li, A.~R. Green-Miller, X.~Hu, A.~Lucic, M.~Mahesh~Mohan, R.~N. Dilger, I.~C. Condotta, B.~Aldridge, J.~M. Hart, and N.~Ahuja, ``Barriers to computer vision applications in pig production facilities,'' {\em Computers and Electronics in Agriculture}, vol.~200, p.~107227, Sept. 2022.

\end{thebibliography}
\bibliographystyle{ieeetr}

}

\medskip
\newpage
\section*{Appendix}
{\textbf{Appendix }\textbf{I}\textbf{ Two types test video clips} } 
\begin{figure}[h]
    \centering
    \includegraphics[width=0.9\linewidth]{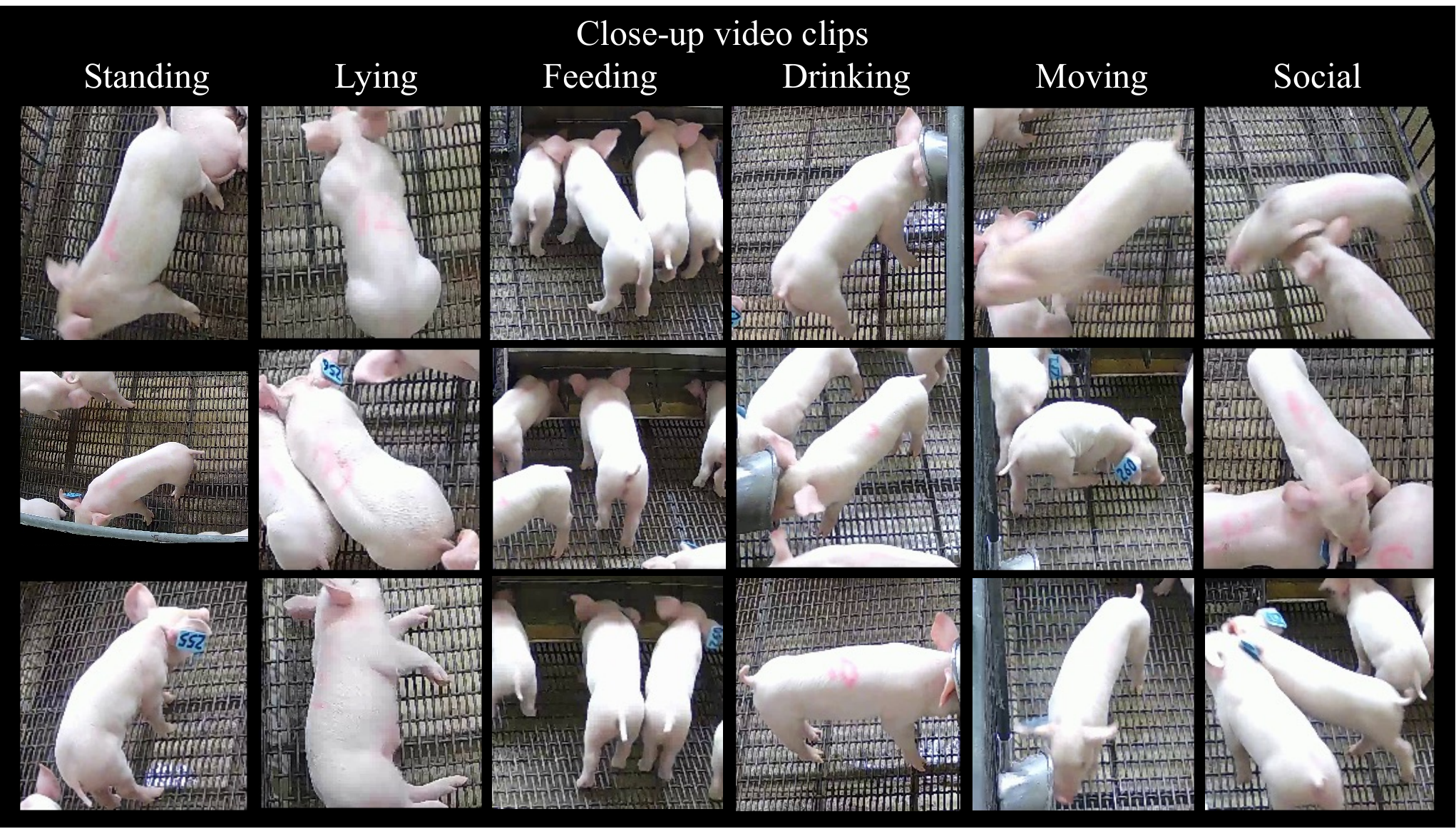}
\end{figure}
\begin{figure}[h]
    \centering
    \includegraphics[width=0.9\linewidth]{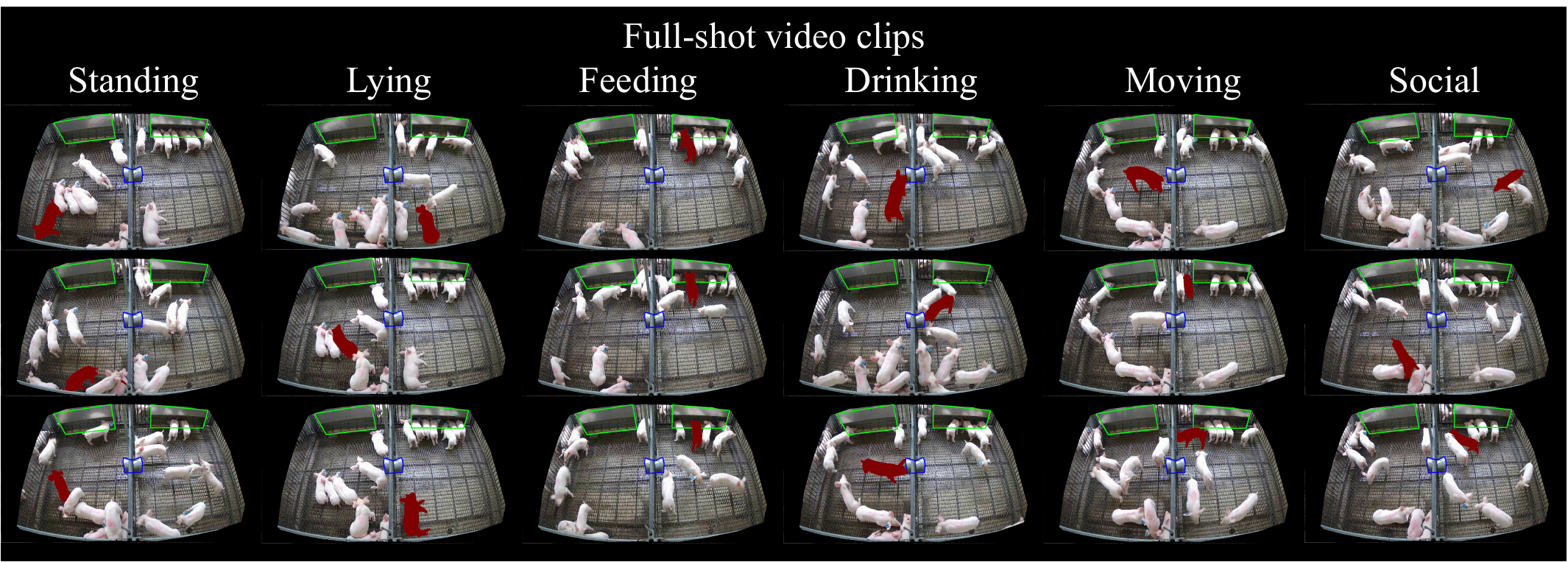}
\end{figure}

{\textbf{Appendix }\textbf{II}\textbf{ Prompt templates}} 
\begin{table}[h]  
    \centering
    \renewcommand{\arraystretch}{1.25}
    \begin{tabular}{p{6cm} p{7cm}}  
        \hline  
        \textbf{A: Text prompt only} & \textbf{B: Text prompt with diverse visual prompting} \\  
        \hline  
        This video depicts a piglet breeding environment. Please observe the video, as the video contains one of the behaviors of the piglet: standing, lying, feeding, drinking, moving, or social behaviors. Your task is to analyze the entire video and identify the type of behavior exhibited by the piglet and the duration of the behavior. If you cannot determine the kind of behavior, please output “Null.” Or, based on your understanding of the pig’s behavior type, and state the duration of the behavior. & This video depicts a piglet breeding environment. The green polygonal area represents the feeding zone, and the blue one represents the drinking zone. First, please determine and answer how many piglets are in this video clip. Among them, the piglet with the red mask is the actor of behavior observation. Analyze the entire video, which contains one of the behaviors of the piglet with the red mask: standing, lying, feeding, drinking, moving, or social behaviors. Identify which category the behavior of the piglet with the red mask falls into and specify the duration of the behavior. If you cannot determine the behavior type, please output “Null.” Or, based on your understanding of the pig’s behavior type, and state the duration of the behavior. \\  
        \hline  
    \end{tabular}  
    
\end{table}

\newpage
{\textbf{Appendix }\textbf{III}\textbf{ Examples of four multimodal LLMs } } 
\begin{figure}[h]
    \centering
    \includegraphics[width=0.75\linewidth]{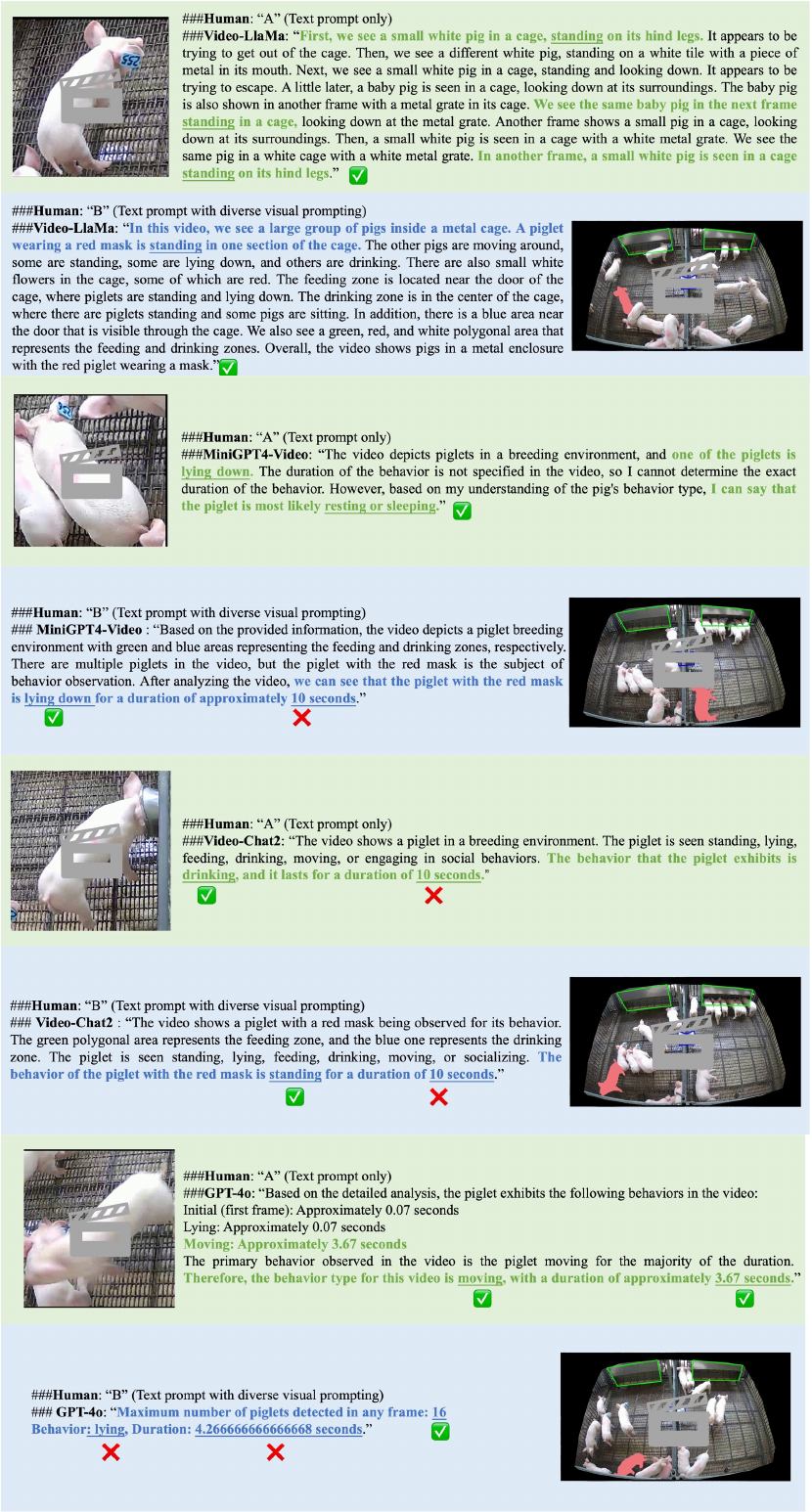}
\end{figure}

\end{document}